\documentstyle[aaai]{article}
\begin{document}

\title{PAL: Pertinence Action Language
}
\author{
Pedro Cabalar~$\!\!^1$, Manuel Cabarcos~$\!\!^1$ and Ram\'on P. Otero~$\!\!^2$\thanks{Visiting from the University of Corunna, Spain.}\\
\\
$^1$ Dept. of Computer Science, \\
University of Corunna, Spain \\
{\tt \{cabalar,cabarcos\}@dc.fi.udc.es}\\
\\
$^2$ University of Texas at El Paso, \\
El Paso, Texas 79968 \\
{\tt otero@cs.utep.edu}
}

\maketitle

\begin{abstract}
\noindent The current document contains a brief description of a system for Reasoning about Actions and Change called PAL (Pertinence Action Language) which makes use of several reasoning properties extracted from a Temporal Expert Systems tool called Medtool.
\end{abstract}

\section{General Information}
% MANDATORY
% 1. On what platforms (e.g., PC, SPARC, ...) does your system work? 
% 2. In what language has it been written? 
% 3. How big is the program? 
The PAL system (standing for Pertinence Action Language) is an interpreter of a causal language for describing action domains and it has been proved in PCs under Linux and Sun workstations under SunOS. It is written in C (using GNU's compiler gcc 2.7.2.1) and sized in 150K of source code, distributed in around 6500 lines. The executable is sized in 65/80K (depending on the platform). The programming effort can be estimated in about 1 man/years.

\section{Description of the System} 

The purpose of PAL is to show the effects of using the concept of {\em Pertinence} (formally introduced in~\cite{Ote99}) in problems of Reasoning about Actions and Change, and to help in establishing semantic features about the logical formalization of this concept. The features of Pertinence have been extracted from a practical tool for Temporal Expert Systems (mainly for the medical domain\cite{TAO,MIM,TKR}) called Medtool~\cite{Medtool} which uses a formalism very close to the current one used in PAL.

The tool is initially intended for solving temporal projection problems, being planning problems reduced to small sized ones, but we expect to use it as a starting point for future efficiency improvements.

\section{Applying the System}

The system can be tested in two different ways, both available in the web page
\begin{verbatim}
http://www.dc.fi.udc.es/ai/~cabalar/pal
\end{verbatim}

The two options are:
\begin{enumerate}
\item For an easy to use way, a simple web-based interface ({\tt PalWeb}) has been implemented. It is enough with writing a domain description in the input box (or just selecting one of the preinstalled examples) and pressing the {\tt Process} button to see the results.
\item The other option is downloading the source code (file {\tt pal1.2.tar.gz}) and compiling it in a local machine. To this aim, follow the next steps:
\begin{verbatim}
gunzip pal1.2.tar.gz
tar -xvf pal1.2.tar
cd pal1.2
make
\end{verbatim}
Finally, for testing the system, try any of the files in the examples directory:
\begin{verbatim}
pal < examples/yale.pal
\end{verbatim}
We may also use the executable in the following way:
\begin{verbatim}
pal examples/yale.pal
\end{verbatim}
\noindent so that standard input remains open to accept new queries or new action executions, providing a rudimentary interpreter.
\end{enumerate}

\subsection{Methodology}

% 1. Is there a general methodology as to how to encode a problem,
%    to take advantage of your work? 
% 2. If there is no general methodology, is it likely to be developed in
%    the near future? 
No general methodology has been developed by now, although due to the closeness to Medtool, many methodological aspects can be directly inherited from practical experience in Medtool expert systems design.

\subsection{Specifics}
                                  
% 1. What is the significance of your system being logic-based? 
% 3. Why is this an important feature of your work? 
The use of logic is the main feature of PAL with respect to previous work done on Medtool expert systems design. The purpose of PAL is to help in the logical formalization for these practical systems so that new formalism features can be incorporated or new tasks like explanation or planning can be done in a coherent way.

As an action formalism, the main features of PAL are:

\begin{enumerate}
\item It is {\em narrative}-based. Each model of a PAL theory has the shape of a finite sequence of situations.
\item It is a {\em causal} formalism. Rules describing the domain behavior are causal rules. Pertinence acts in a similar way to Lin's $Caused$ predicate or to {\em occlusion}, although there are fundamental differences (see~\cite{Ote99}), especially on how rule conditions are interpreted.
\item It allows {\em concurrent actions}.
\item Actions and fluents are {\em functional}, though limited to finite domains and codomains. Arithmetic expressions relating actions and fluents are allowed.
\item It allows both ``real'' execution and hypothetical reasoning. There exists a real narrative which can be updated in an execution trace and whose (current and past) results can be consulted at any time. Hypothetical reasoning allows to test the results of an hypothetical execution, or to find plans that satisfy a given expression for a future situation.
\end{enumerate}

% 2. What is the semantics of your system (e.g. propositional logic,
%    first-order logic, default logic, stable or well-founded semantics, etc.)? 

Each high level domain description is translated into a set of ground rules using a propositional representation with two kinds of atoms: $holds(Q,value,situation)$ and $pertinent(Q,situation)$ (being $Q$ a fluent or an action). The system is intended for testing different semantics for these ground rules. To this aim, the rules behavior has been implemented in a separated module so that it can be easily changed. The first approach we have done is a version based on Well Founded Semantics, since it is the closest one to the way in which the practical tool currently works. This well founded version interprets the set of rules as an objective logic program, leaving default negation just for minimizing $pertinent$ atoms. The iterative algorithm for computing the well founded model has been specialized so that pertinence minimization and inertia axioms are, in fact, executed implicitly. 

A second option that will be soon available is using as ``semantics'' the inference mechanism of Medtool (which was already implemented in a C library known as Medtool-Connection). Of course, this would be an operational or procedural solution, but it is useful for comparing the practical tool results with the different logic-based versions. We also expect to incorporate a future Stable Models version (calling to smodels) and a Completion based version (using some SAT prover, following a similar technique to CCALC).

% 4. How will your work influence other areas such as artificial
%    intelligence,  knowlege bases and deductive databases? 

The work could influence both areas of Reasoning about Actions and Temporal Expert Systems. The interest for the former could be theoretical (the introduction of pertinence in action domains) but also practical, since PAL formalization could eventually allow studying Medtool expert systems as action theories. The interest for Temporal Expert Systems is that PAL provides an underlying logical formalization. Other less related areas could also be influenced. For instance, a strong relationship between Medtool practical formalism and Systems Theory formalizations like DEVS (Discrete Events Systems Specification) has been established (\cite{Otero94,Otero96}). Finally, we are also studying the relation between the use of the real narrative done in PAL and approaches based on Temporal Deductive Databases.

\subsection{Users and Useability} 
% 1. Must the potential user be an expert in logic or need to have a 
%    deep knowledge of the specific area of the application?

In order to use the high level language, the user does not need to be an expert in logic, although some notions about pertinence and logic-based systems is recommendable. The basic syntax is quite direct, following the style of Medtool formalism, on which PAL is inspired. In fact, Medtool formalism is currently being used by the experts (mainly physicians), so that they directly develop the expert systems, under some minimal assistance. However, in order to use PAL as a comparison tool for different semantic implementations, a deep knowledge about Nonmonotonic Reasoning and Logic Programming semantics will be required.

% 2. Is your system sufficiently flexible to handle other problems? 
% 2. If so, what kind of problems, and what would have to be done
%    to apply your system? 

As an example of flexibility, although it is not one of the initial purposes, PAL can also be indirectly used as a Logic Programming tool under Well Founded semantics. An example which includes the transformation steps to this aim can be found in PAL distribution (file {\tt wf.pal}).

% 3. Is your system applied outside your research group or even in other areas? 
% 4. Is it likely to be used in the future (are there concrete plans to do so)? 
The PAL system is currently under continuous development and has only been used by our research group until now. However, Medtool formalism, on which PAL is inspired, has been used for constructing several real Expert Systems\cite{TAO,MIM,TKR}, but also in the Financial domain.

\section{Evaluating the System}

\subsection{Benchmarks}
% 1. How can users evaluate comparative systems?
% 2. For example, are there accepted benchmarks for testing your 
%    system and comparing it to others? 

Due to the premature stage of PAL, little comparative evaluation work has been done yet. The kind of properties usually evaluated in actions formalizations are more theoretical, especially related to flexibility and elaboration tolerance, than practical, like memory usage or response time, although several approaches are already obtaining successful results for these last parameters. As explained in the description section, many usual action topics like, for instance, qualifications, defeasible rules or delayed effects, are not covered yet by PAL,  although their introduction is a subject of current research. However, we think that the current version provides enough expressivity for a practical use, properly dealing, for instance, with the frame and ramification problems (as shown by standard examples included in files {\tt yale.pal} and {\tt suitcase.pal}).

Unfortunately, there are not too many benchmarks for testing action approaches yet. As an exception, we could mention the recent initiative of the Logic Modelling Workshop\footnote{{\tt http://www.ida.liu.se/ext/etai/lmw/}}, whose examples will be studied under PAL formalization. 

As for PAL time performance, we could mention that the system works acceptably well for temporal projection problems. For instance, the execution of 5 transitions for the domain example {\tt counter.pal}, which contains a causal chain of 1000 fluents, is performed in 0.85 seconds\footnote{All times obtained in a 133MHz Pentium with 32M of RAM and under Linux Slackware 2.0.29.}. Also, some small planning problems can be performed in an reasonable way. For instance, it takes 0.42 seconds for finding all the 781 ways for opening Lin's suitcase in 5 situations (allowing concurrent actions and no execution of any action). Planning may become excessively hard, since it is attached in a naive way, generating all the possible combinations of actions and allowing concurrent execution. A nonconcurrent actions option has been added for better planning performance in most usual domains. Thus, for instance, a solution of 11 situations long for the missionaries and cannibals problem is found in 0.13 seconds, and the 4 solutions for that length are obtained in 0.7 seconds.

% 3. How user friendly is the system? 
% 4. What are important features for such systems? 

With respect to user-friendliness, the main advantage is the comfortability of domain descriptions involving numerical variables (limited to finite and discrete domains). For instance, the missionaries and cannibals problem can be described using 9 high level rules. With respect to applying the system, the current version is mainly batch-oriented, although it can be used as a rudimentary interpreter by entering sentences from standard input. We plan to extend this capability to a real interpreter interface. 

\subsection{Comparison} 
% 1. Can your system compete with other special-purpose, 
%    possibly non logic-based systems? 
% 2. If yes, how exactly would you be able to compare it to such systems? 
% 3. Are there generally accepted benchmarks for these areas? 

As many action approaches, the PAL system can compete with planning systems, particularly in the expressivity of their representational formalism. However, no serious comparison to usual planning benchmarks has been established yet.

\subsection{Problem Size} 
% 1. What is the current problem size your system can handle? 
Temporal projection domains can be reasonably handled for thousands of fluents. Planning problems, however, are quite limited yet, specially when dealing with a small/medium number of situations and concurrent execution of actions.

% 2. Is your system a prototype, or do you believe it
%    can be used on problems of realistic size? 
% 3. Assuming your system is a prototype, what has to be done to 
%    assure that it scales to large-sized problems? 

The system must be considered as a prototype, but several efficiency improvements (goal-directed planning, taking benefit of particular structures in the rules, heuristics, etc) are currently under study and could allow handling larger planning problems.

\section{An example: the blocks world}

In this last section, we include, as a piece of syntax example, a possible representation of the blocks world domain. We will not provide an exhaustive description of PAL syntax, but will emphasize instead those most relevant features of the formalism. A PAL domain description consists of two parts: declarations and sentences. In the declarations part, we define the sets of fluents, actions and rules, using perhaps some auxiliary definitions of constants and sets to this aim. For instance, we will declare the following sets:

\begin{verbatim}
sets
  block = [1,4];
  location = block + {table};
\end{verbatim}

Intuitively, these lines define the set of blocks as the integers from 1 to 4, and the set of locations as any block or the table. A set is usually defined as a group of elements embraced by {\tt \{\dots \} }, like for instance the singleton set {\tt \{table\}}. Elements in a set can be both symbol names (an identifier with a lower-case initial, like {\tt table}) or integer numbers. An alternative way of defining a set is using the interval notation, so that {\tt [1,4]} is equivalent to {\tt \{1,2,3,4\}}. Set expressions can be constructed using binary operators {\tt +,-,*} standing for union, difference and intersection respectively.

Actions and fluents are considered to be functions. Thus, an action or fluent definition consists of the descriptions of its domain and codomain using the standard mathematical notation:

\begin{verbatim}
f: set1 x set2 x set3 -> set4
\end{verbatim}

Note that, although {\tt x} is used here as an operator, it can also be used as an identifier, depending on the grammar context. As an example of functional action, we define:

\begin{verbatim}
actions
  carry: block -> location;
\end{verbatim}

\noindent so that we carry a block to a unique location. Notice that, in this way, we implicitly specify that a block cannot be placed on two different locations, since the value of a function (i.e. each {\tt carry(B)}) is {\em unique}. In the example, we will use the following fluents:

\begin{verbatim}
fluents
  loc: block -> location;
  free: block -> {true,false};
\end{verbatim}

When a fluent is boolean, we can simply ignore the codomain definition in the following way:

\begin{verbatim}
  free: block;
\end{verbatim}

In the same way, we could define actions or fluents without domain. Imagine an action for just marking a unique block:

\begin{verbatim}
actions
  mark: -> block;
\end{verbatim}

\noindent or a fluent that points if we have made some mark:

\begin{verbatim}
fluents
  markdone: -> {true,false};
\end{verbatim}

This last case can be further abbreviated as:

\begin{verbatim}
fluents
  markdone;
\end{verbatim}

In the declarations section we can also define variables. In PAL, variables are identified by an upper-case initial, and vary on a given sort. They are used for defining rule schemata and for making complex queries. We will use the following two block variables:

\begin{verbatim}
vars
  B,C : block;
\end{verbatim}

Finally, the declarations section would contain the set of rules describing the system behavior:

\begin{verbatim}
rules
  loc(B):=carry(B);
  not free(C) if carry(B)=C;
  free(B) if pert(carry(C)) and prev(loc(C))=B;
  false if pert(carry(B)) and not prev(free(B));
  false if carry(B)=C and not prev(free(C));
\end{verbatim}

Rules have a head of shape:
\begin{verbatim}
<fluent>:=expr
\end{verbatim}
and an optional condition preceeded by the conditional operator {\tt if}. In a rule head we allow the abbreviations:
\begin{verbatim}
<fluent>
not <fluent>
\end{verbatim}
\noindent standing for:
\begin{verbatim}
<fluent>:=true
<fluent>:=false
\end{verbatim}
\noindent respectively, and we allow constraint rules with a {\tt false} head. 

Coming back to our example, the first rule says that whatever the block {\tt B} we carry, its current location {\tt loc(B)} will be the value given by the {\tt carry(B)} action. The second rule asserts that a block {\tt C} becomes not free when we carry some {\tt B} on top of it. The third rule says that if {\tt carry(C)} is {\em pertinent}, that is, we carry {\tt C} without regarding to which location, and the previous location of {\tt C} was {\tt B} then {\tt B} becomes free. The fourth rule asserts that we cannot perform a {\tt carry(B)} when {\tt B} was not free, whereas the fifth one, avoids carrying {\tt B} to a nonfree block {\tt C}.

The sentences part may contain declarations about the actual narrative or queries. The actual narrative is first specified by providing the initial situation:

\begin{verbatim}
initially
  loc(B):=table,free(B);
\end{verbatim}

This means that all the blocks are on the table and are free. Performing actions in the actual narrative can be done in the following way:

\begin{verbatim}
do {carry(1):=2;}
\end{verbatim}

This would carry block 1 on top of block 2. The output would be the following one:

\begin{verbatim}
1)
carry(1):=2
loc(1):=2
free(2):=false
\end{verbatim}

The output shows only the pertinent facts, just pointing out the performed action and its derived effects (all the rest has persisted). We can perform several actions in a sequence:

\begin{verbatim}
do {carry(1):=table;carry(2):=3; carry(1):=2;}
\end{verbatim}

\noindent which would have as output:

\begin{verbatim}
2)
carry(1):=table
loc(1):=table
free(2):=true
3)
carry(2):=3
loc(2):=3
free(3):=false
4)
carry(1):=2
loc(1):=2
free(2):=false
\end{verbatim}

Notice how performing actions is an incremental task: the execution of the sequence generates situations from 2 to 4, since it was performed taking 1 as the initial situation (that is, the one we had obtained previously). We can reinitialize (``resume'') the actual narrative by providing a new {\tt initially} clause. Besides, actions can also be performed concurrently, or we can even perform no action:

\begin{verbatim}
initially
  loc(B):=table,free(B);
do {carry(1):=2,carry(3):=4; carry(1):=3; ; }
\end{verbatim}

\noindent that is, we first simultaneously move 1 on top of 2 and 3 on top of 4, in the next situation, we move 1 to 3, and finally we perform no action. The output would be:
\begin{verbatim}
Resume
1)
carry(1):=2
carry(3):=4
loc(1):=2
loc(3):=4
free(2):=false
free(4):=false
2)
carry(1):=3
loc(1):=3
free(2):=true
free(3):=false
3)
\end{verbatim}

Finally, we will briefly comment the other kind of sentences: queries. The simplest kind of query allows asking about the current state. For instance, we can check that block 2 is free and it is on table:

\begin{verbatim}
query
  free(2) and loc(2)=table?
  
yes  
\end{verbatim}

Variables can be used in a similar way to Prolog queries. For instance, we can find all the blocks that are not free:
\begin{verbatim}
query
  not free(B)?
  
B=3
B=4  

2 solutions
\end{verbatim}

The general shape of queries allows asking properties about an hypothetical future. Answers will have the shape of plans, that is, sequences of actions to be performed for satisfying the query. For making some tests, we will add the assumption of non-concurrency of actions (otherwise, we should add a rule for specifying that there is not space for two blocks on top of a third block). The  non-concurrency assumption is specified in the declarations part, for instance, as:

\begin{verbatim}
options
  not concurrent;
\end{verbatim}

Non-concurrency also assumes that an (unique) action is {\em mandatorily} performed. Now, using the initial state:

\begin{verbatim}
initially
  loc(B):=table,free(B);
\end{verbatim}

\noindent we can try to make block 3 not free in the next situation:

\begin{verbatim}
query
  true;not free(3) ?

\noindent obtaining 4 possible answers:

\begin{verbatim}
Resume

Solution 1:
1)
carry(1):=3
loc(1):=3
free(3):=false

Solution 2:
1)
carry(2):=3
loc(2):=3
free(3):=false

Solution 3:
1)
carry(3):=3
loc(3):=3
free(3):=false

Solution 4:
1)
carry(4):=3
loc(4):=3
free(3):=false

4 solutions
\end{verbatim}

Notice that in the current (present) situation we do not require anything and we use the {\tt true} expression. When this happens, we can abbreviate it as an empty expression like in:

\begin{verbatim}
query ; ; ; not free(3)?
\end{verbatim}
\noindent that would look for the ways of getting block 3 not free after 3 steps (this generates 1072 solutions!). We can fix the number of solutions we wish by another option:

\begin{verbatim}
options
  solutions=1;
\end{verbatim}

It is also possible to repeat the same expression a number of times along several situations. For instance, we can ask the queries:

\begin{verbatim}
query 
  ...{3} not free(3) ?
  
  free(1) ...{3} not free(3)?
\end{verbatim}

\noindent so that the first one replaces our immediately previous example, and the second one tries to find ways of occupying block 3 in 3 steps but maintaining block 1 free. 

\subsubsection{Acknowledgments.}

%We want to thank to ...
This research is partially supported by the Government of Spain, grant PB97-0228.

\bibliographystyle{aaai}
\bibliography{nmr00}

\end{document}